# Construction of a classification model for dementia among Brazilian adults aged 50 and over

Authors: Menezes, FS[1,3]; Barretto, MCFG[2,3], Garcia, EQC[2,3], Ferreira, TAE[2], Alvez, JG[1]

[1] Instituto de Medicina Integral Professor Fernando Figueira (IMIP)

[2] Universidade Federal Rural de Pernambuco (UFRPE)

[3] Future Tech

**\* Correspondence:**
Corresponding Author
felipemenezes.contato@gmail.com

## Abstract

Background: Dementia is a multifactorial and debilitating condition marked by cognitive decline and behavioral changes that compromise independence and daily activities. This condition is a growing challenge in low- and middle-income countries such as Brazil, and early identification of associated factors can guide preventive strategies and health policies. Objectives: To build a dementia classification model for middle-aged and elderly Brazilians, implemented in Python, combining variable selection and multivariable analysis, using low-cost variables with modification potential. Methods: Observational study with a predictive modeling approach using a cross-sectional design, aimed at estimating the chances of developing dementia, using data from the Brazilian Longitudinal Study of Aging (ELSI-Brazil), involving 9,412 participants. Dementia was determined based on neuropsychological assessment and informant-based cognitive function. Analyses were performed with Random Forest (RF) and multivariable logistic regression to estimate the chances of dementia in the middle-aged and elderly population of Brazil. Results: The prevalence of dementia was 9.6%. The highest odds of dementia were observed in illiterate individuals (Odds Ratio (OR) = 7.42; 95% Confidence Interval (CI): 4.04-13.62), individuals aged 90 years or older (OR = 11.00; 95% CI: 5.05-23.95), low weight (OR = 2.11; 95% CI: 1.12-3.97), low handgrip strength (OR = 2.50; 95% CI: 1.09-5.76), self-reported black skin color (OR = 1.47; 95% CI: 1.07-2.00), physical inactivity (OR = 1.61; 95% CI: 1.25-2.08), self-reported hearing loss (OR = 1.65; 95% CI: 1.16-2.37), and presence of depressive symptoms (OR = 1.72; 95% CI: 1.36-2.16). In contrast, higher education (OR=0.44; 95% CI: 0.21-0.94), greater life satisfaction (OR=0.72; 95% CI: 0.52-0.99), and being employed (OR=0.78; 95% CI: 0.61-1.00) were protective factors. The RF model outperformed logistic regression, achieving an area under the ROC curve of 0.776 (95% CI: 0.740–0.811), with sensitivity of 0.708, specificity of 0.702, F1-score of 0.311, G-means of 0.705, and accuracy of 0.703. Conclusion: The findings reinforce the multidimensional nature of dementia and the importance of accessible factors for identifying vulnerable individuals. Strengthening public policies focused on promoting brain health can contribute significantly to the efficient allocation of resources in primary care and dementia prevention in Brazil.

## 1 Introduction

Dementia is defined by the World Health Organization (WHO) (1) as a progressive neurological condition characterized by impairment in multiple cognitive functions, including memory, reasoning, language, and the ability to perform everyday activities, exceeding the decline expected from normal aging. It is a clinical syndrome that may be triggered by various neurodegenerative pathologies, ultimately leading to the gradual and irreversible loss of nerve cells and to structural and functional brain impairment. Cognitive deterioration is often accompanied by behavioral, emotional, and motivational changes, which substantially affect individual autonomy and quality of life. The impact of dementia extends beyond the affected individual, generating physical, emotional, social, and economic consequences for families, caregivers, and healthcare systems.

In Brazil, the Ministry of Health estimates that approximately two million Brazilians are living with some form of dementia. However, preliminary data indicate that over 70% of those affected remain undiagnosed, which further heightens concern (2), especially considering that recent estimates show an overall dementia prevalence of 5.8% among people aged 60 and older. Prevalence is higher among women (6.8%) than men (4.6%), ranging from 3.2% in the 60–64 age group to 42.8% among individuals aged 90 and above, and is more prevalent among illiterate individuals (16.5%) compared to those with a university-level education or higher (2.1%) (3).

A transitional state between normal cognitive functioning and dementia is known as cognitive impairment, which is characterized by a decline greater than expected for a person's age and educational level. It affects memory, attention, language, and reasoning abilities, but does not significantly interfere with daily activities (4). Estimates indicate a prevalence of 8.1% among individuals aged 60 and older. As with dementia, cognitive impairment is less common in men (6.8%) than in women (9.1%). Its prevalence remains stable across age groups from 60 to 79 but increases to 11.8% and 12% in the 80-84 and 85-89 age groups, respectively. Interestingly, it decreases to 10% among those aged 90 and over. Regarding educational level, prevalence ranges from 9.4% among illiterate individuals to 10.8% among those with a university-level education or higher (3).

In recent years, efforts have been made to identify potential risk factors and early intervention strategies aimed at reducing the likelihood of dementia onset and delaying its progression. It is plausible to assume that anticipating action regarding certain modifiable risk factors could postpone the onset of the disease, given that dementia has prodromal characteristics and may begin to manifest decades before clinical symptoms appear (5). In a global context, it is vital to consider the significant rise in dementia incidence in Low- and Middle-Income Countries (LMICs) compared to High-Income Countries (HICs) (6). The early and accurate identification of individuals at high risk for dementia plays a crucial role in the effective implementation of preventive strategies.

It is imperative to have an assessment that is clinically feasible for identifying individuals at high risk, especially considering that in Latin American countries such as Brazil, there are significant challenges related to dementia development indicators (6). These challenges are reflected in key characteristics, such as: (i) low income; and (ii) high levels of social inequality (6). These socioeconomic factors often pose barriers to accessing high-cost diagnostic tools such as neuroimaging or biomarkers, which directly affect early detection and effective monitoring. In light of these limitations, there is a pressing need to consider alternative approaches, using accessible variables for both clinical use and large-scale community-based studies.

The current epidemiological context demands a strong prioritization of dementia across all levels, from local settings to the global stage (7). Identifying modifiable factors that can be addressed in advance helps to promote brain health, especially considering that approximately 45% of dementia cases could be delayed or prevented through risk factor modification (6). This underscores the urgent need for innovation and research in monitoring these variables (8,9). Evidence shows that machine learning algorithms outperform traditional models in predictive performance within medical contexts, due to their ability to manage complex and non-linear relationships. However, traditional models offer advantages such as transparency and interpretability, which are essential in clinical research settings (10–12). In this regard, combining both approaches allows us to leverage the strengths of each when it comes to predicting health outcomes.

Currently, a number of models have been developed for predicting cognitive impairment and dementia (13–16), particularly using accessible variables that encompass demographic, socioeconomic, cognitive, functional, and physical factors (17). However, it is important to highlight that all of these models were developed using data from populations in foreign countries, which may present a significant limitation when generalizing findings to the Brazilian population. This underscores the need for comparative studies or detailed analyses to assess their effectiveness and relevance within the Brazilian context. Furthermore, a Brazilian study (18) developed a predictive model for cognitive impairment based on accessible variables in primary care, using a machine learning approach. While this contribution is significant, it is important to note that the data used are not representative of the broader middle-aged and older adult population in Brazil, as the sample was composed mainly of civil servants, who generally have higher levels of education and income compared to the national average. This limitation may affect the generalizability of the results to the broader Brazilian population, particularly in relation to social inequalities and disparities in access to healthcare.

Given this scenario, there is a clear need to develop predictive models for dementia using data that are representative of the Brazilian middle-aged and older adult population, specifically taking into account sociodemographic diversity and challenges related to access to high-cost diagnostic exams. The use of accessible, low-cost variables combined with innovative approaches – such as machine learning integrated with traditional statistical analysis – offers a meaningful contribution to the early identification of individuals at risk, as well as to more effective and targeted interventions in primary care settings.

In this context, the present study aims to address a gap in the national literature by building a dementia classification model for Brazilian middle-aged and older adults. The model is implemented in Python and combines variable selection (Random Forest algorithm) with multivariable analysis (Logistic Regression), analyzing low-cost and potentially modifiable variables associated with the outcome (adjusted odds ratio).

## 2      Materials and Methods

### 2.1 Study Characteristics

This is an observational, cross-sectional study with a predictive modeling approach aimed at estimating the likelihood of developing dementia. The Brazilian Longitudinal Study of Aging (ELSI-Brazil) was funded by the Brazilian Ministry of Health: DECIT/SCTIE – Department of Science and Technology of the Secretariat of Science, Technology and Strategic Inputs (Grants: 404965/2012-1 and TED 28/2017); and COPID/DECIV/SAPS – Coordination of Health for the Older Adult in Primary Care, Department of Life Cycles of the Secretariat of Primary Health Care (Grants: 20836, 22566, 23700, 25560, 25552, and 27510). Ethical approval was granted by the FIOCRUZ Ethics Committee in Minas Gerais, and the study was registered on the *Plataforma Brasil* (CAAE: 34649814.3.0000.5091). All participants signed an informed consent form prior to data collection. This study was also approved by the Research Ethics Committee of the Instituto de Medicina Integral Professor Fernando Figueira (CEP-IMIP), under CAAE: 88342525.5.0000.5201. The study was conducted in accordance with the principles of the Declaration of Helsinki, the guidelines of Resolution 466/12 of the Brazilian National Health Council, and the General Data Protection Law (Law No. 13.709, of August 14, 2018).

## 2.2 Eligibility Criteria and Data Collection Procedure

### 2.2.1 Inclusion Criteria

- All individuals aged 50 years or older, non-institutionalized, residing in Brazil, who participated in the individual interview during the first wave.

### 2.2.2 Exclusion Criteria

- Incomplete data from the cognitive test battery or informant-reported cognitive function;
- Variables are not relevant to the objectives of the study.

## 2.3 Data Collection and Study Population

The data were obtained from the ELSI-Brazil database, a population-based cohort study with a meticulously planned sampling strategy designed to accurately reflect the diversity of the non-institutionalized population aged 50 years and older. The sample includes individuals from both urban and rural areas across municipalities of varying sizes in Brazil, with data collection conducted between 2015 and 2016 (19,20). The final sample consisted of 9,412 participants residing in 70 municipalities from different geographic regions. A complex, multi-stage sampling design was used, involving stratification of primary sampling units (municipalities), census tracts, and households. Municipalities were grouped into four strata based on population size. In the first three strata, sampling was conducted in three stages: selection of municipalities, census tracts, and households. In the fourth stratum, which included the largest municipalities, sampling was conducted in two stages: selection of census tracts and households (19,20). All residents aged 50 years or older in the selected households were eligible for interviews and physical measurements.

## 2.4 Research Tools

### 2.4.1 Cognitive Function

For the dependent variable, a participant classification approach was used to identify cognitive status within the sample, distinguishing between normal cognition, cognitive impairment, and dementia. This classification was based on a combination of cognitive performance and functional capacity in Instrumental Activities of Daily Living (IADLs) (3).

To assess cognitive performance, a battery of neuropsychological tests was used to evaluate different cognitive domains, namely: temporal orientation, semantic verbal fluency, a 10-word list test for immediate and delayed recall, prospective memory, and semantic memory. A detailed description of all subdomains of the neuropsychological test battery can be found in Supplementary Table 1.

Initially, the numerical results from each test subdomain were used to generate raw scores, such as the number of correct answers and/or performance rankings on the test (211). For the questions related to temporal orientation (day, month, year, and day of the week) and semantic memory (naming certain objects and political figures), one point was assigned for each correct answer. These were then summed to produce a raw score for each participant. For the subdomains of semantic verbal fluency and the 10-word list for immediate and delayed recall, the number of words (animals) mentioned, and the number of words recalled (immediately and after a delay), respectively, were recorded. Prospective memory was assessed using a scoring system ranging from 0 to 5, with the correct execution of the requested task receiving the highest score.

Next, raw scores were standardized using z-score calculation, which involved subtracting each individual's score from the sample mean and dividing the result by the sample standard deviation. This yielded a standardized score for each participant, allowing the results to reflect deviations from the mean and to be comparable across different subdomains. The meaning of the standardized scores from all subdomains was then used to compute a global cognitive score.

**2.4.2 Informant-Based Cognitive Function**

For participants unable to complete the cognitive test battery, the short version of the *Informant Questionnaire on Cognitive Decline in the Elderly* (IQCODE), consisting of 16 items and validated in Brazil (22-24), was administered. Based on the results, individuals were classified as having normal cognition, cognitive impairment, or dementia.

The IQCODE is the instrument used by ELSI-Brazil to assess cognitive function based on information provided by someone close to the participant, such as a family member or caregiver. It aims to identify cognitive changes, particularly in cases where the individual may be unable to accurately report their own condition.

This section (*Cognitive function for a proxy respondent*) consisted of questions directed at the informant, who was asked to compare the participant's current memory and cognitive abilities with their condition two years prior. The questions were structured to address various aspects of memory and cognition, with responses categorized into pre-established options. The informant was instructed to respond based on their perception of the degree of improvement, stability, or decline.

Responses were collected using standardized options such as "Improved," "Did not change much," or "Worsened," along with subcategories to indicate the intensity of the changes, such as "Improved a lot," "Some improvement," "Some decline," or "Worsened a lot." The options "Don't know/Did not respond" were used in cases where the informant lacked sufficient information to answer, and these were treated as missing data and excluded from the analysis. All 16 items of the short version of the IQCODE are detailed in Supplementary Table 2.

Each of the 16 IQCODE items reflects the degree of perceived change, rated using a Likert scale as follows: (1) improved a lot, (2) some improvement, (3) did not change much, (4) some decline, and (5) worsened a lot. The final score is calculated as the average of all valid responses – i.e., only the items that were answered – allowing for the handling of missing data. Participants with scores below 3.22 are categorized as having normal cognition, as this indicates no significant cognitive decline. Those scoring between 3.22 and 3.47 are considered to have cognitive impairment, suggesting a mild perceived decline not sufficient to be classified as dementia. Lastly, participants with scores equal to or greater than 3.48 are classified as having dementia, reflecting a higher degree of cognitive deterioration as perceived by the informant (22,25).

**2.4.3 Neuropsychological Test Norms**

To calculate the z-scores that comprise the global cognitive score, it is first necessary to establish correction factors or norms based on the raw scores of a normative subsample. The purpose is to provide regression-based adjusted norms (26) for evaluating the cognitive performance of the full sample. This normative subsample consists of "healthy" individuals without diseases or conditions that could lead to cognitive or behavioral impairments affecting performance on the cognitive test battery.

Full details of each variable used to define the normative subsample are provided in Supplementary Table 3. Below is a brief description of the excluded variables:

- Self-reported visual and hearing impairments;
- Self-reported medical diagnosis of depression or depressive symptoms assessed using the eight-item Center for Epidemiologic Studies Depression Scale (CES-D8), with a cutoff score of 4 (27,28);
- Self-reported history of diagnosis of stroke, Alzheimer's disease, or Parkinson's disease;
- Excessive alcohol consumption based on the criteria of the National Institute on Alcohol Abuse and Alcoholism (NIAAA), defined as weekly intake of 14 drinks or daily intake of 4 drinks for men, and weekly intake of 7 drinks or daily intake of 3 drinks for women (29–31);
- Self-reported memory complaints or those reported by an informant;
- Self-reported impairment in 4 instrumental activities of daily living, such as managing one's own finances and medications, using transportation, and using the telephone;
- Missing data and absence of cognitive test results.

A multiple regression analysis was performed using the standardized global cognitive score as the dependent variable and age, sex/gender, and education (in years) as predictors. Based on the weighted coefficients obtained from the regression model, predicted global cognitive scores were calculated for the entire sample. These predicted scores were then subtracted from the actual scores of each participant, generating regression residuals. The residuals were standardized by dividing them by the standard deviation of the residuals (root mean square error) from the regression performed on the normative subsample. The result was a standardized z-score for each participant's global cognitive score, which was used to determine the presence of cognitive impairment according to previously established criteria.

**2.4.4 Categorization of the Response Variable**

A classification of the participants in the sample was carried out to identify individuals with normal cognition, cognitive impairment, or dementia, based on the evaluation of cognitive function previously described, in combination with assessments of functional capacity. This approach was employed in a prior study using the same database (ELSI-Brazil) (3).

According to the regression-based normative standards, a global cognitive score (z-score) equal to or lower than -1.5 (one and a half standard deviations below the mean) was considered indicative of cognitive impairment. Based on these results, it was possible to determine each individual's cognitive status, considering both cognitive performance (through the global cognitive z-score) and functional impairment, as assessed through IADLs. Functional impairment was defined by self-reported difficulty in at least four IADLs that are directly related to cognitive function, namely: managing finances, using transportation, using the telephone, and managing medications.

The classification followed these criteria: 1) Normal cognition was defined as the absence of both cognitive and functional impairment, or the presence of functional impairment not related to cognitive issues (e.g., due to physical limitations), along with an IQCODE score below 3.22 (22,25); 2) Cognitive impairment was defined as cognitive impairment without functional impairment, or an IQCODE score equal to or greater than 3.22 (22,25); 3) Dementia was defined as the presence of both cognitive and functional impairment, or an IQCODE score equal to or greater than 3.48 (22,25), for participants who were unable to complete the cognitive assessment and required evaluation via the IQCODE.

**2.4.4.1 Sociodemographic Characteristics**

In this study, we examined six sociodemographic factors: 1) Age – individuals aged 50 years or older, based on their age at the time of the interview, categorized into 5-year age groups; 2) Sex – male and female; 3) Educational level – defined by the highest school grade the individual had successfully completed (never attended school; 1st to 8th grade of primary education; 1st to 3rd grade of secondary education; adult education/equivalency programs and some college; completed higher education; and postgraduate degrees such as specialization/residency, master's, or doctoral degrees); 4) Skin color – white, Black, Brown (mixed race), Yellow (East Asian origin, such as Japanese, Chinese, Korean, etc.), and Indigenous; 5) Occupational status in the past 30 days; 6) Marital status – single, married/living with partner/in a stable union, divorced or separated, and widowed.

**2.4.4.2 Health and Comorbidities**

Body Mass Index (BMI) was calculated as the ratio between body weight in kilograms (kg) and height in meters squared (m²), using the formula: $BMI = weight(kg)/height(m)^2$. To ensure accuracy, measurements were taken with individuals barefoot, standing upright with feet and heels together, and with their back and head against the measuring device. Measurements were taken twice, and the average value was used (32). In accordance with World Health Organization (WHO) guidelines (33), BMI was categorized to facilitate data analysis. Although BMI does not directly measure body fat percentage and may overestimate it in highly muscular individuals or underestimate it in those with higher fat mass (34), it remains widely used for being simple, standardized, low-cost, and quick making it particularly useful in population-based and international studies (6,35).

Blood pressure (BP) was categorized using a cutoff of ≥140/90 mm Hg, based on the Brazilian guidelines for hypertension (36), as adopted in previous studies (37). BP was measured using a validated automatic device (38), the Omron M3 HEM-7200 (Omron Healthcare Brazil, São Paulo, Brazil). Before measurement, participants were instructed to remain seated and still for at least five minutes. During this time, they were asked not to speak or move, ensuring their blood pressure reached a resting level suitable for accurate measurement. Three measurements were taken, with two-minute intervals between each, and the average BP was calculated (32). BP was then categorized into hypertensive or non-hypertensive.

Diabetes was self-reported and identified using the question: "Has a doctor ever told you that you have diabetes (high blood sugar)?" A "yes" response classified the individual as having diabetes. This self-report approach is widely accepted and has been used in several previous population-based studies (39–41), supporting its validity and reliability. Additionally, other studies have confirmed that self-reported diabetes is a significant predictor of mortality risk among older adults (42).

High cholesterol was self-reported through the question: "Has a doctor ever told you that you have high cholesterol?" with binary response options ("yes" or "no"). Self-reported high cholesterol emerged as the third most common condition among participants, affecting 30.5% of the sample (43). Including high cholesterol as a predictor variable is highly relevant. Its high prevalence among older adults not only reflects a major public health issue, but may also influence the management of other chronic conditions (44,45). Furthermore, recent evidence supports its role as a potentially modifiable risk factor for dementia (6).

Untreated vision loss due to conditions such as cataracts and diabetic retinopathy was collected via self-report in the ELSI-Brazil study. Two questions were used to define cataract status: "Has an ophthalmologist ever told you that you have or had cataracts in one or both eyes?" and "Have you had cataract surgery?" These questions were selected based on evidence showing that individuals with cataracts are at increased risk for all-cause dementia (46), and that cataract removal is associated with significantly reduced dementia risk (47). In our sample, those who reported a cataract diagnosis but had not undergone surgery were classified as "with cataracts," while those who had never been diagnosed or who had undergone surgery were classified as "without cataracts."

Regarding diabetic retinopathy, we used the question: "Has an ophthalmologist ever told you that you have or had diabetic retinopathy (diabetes in the eyes)?" since this condition has been associated with an increased risk of dementia (45,48). All questions had binary response options ("yes" or "no"). Responses such as "don't know/did not answer" were treated as missing data. Including these predictor variables may enrich the analysis, as they have emerged as significant and potentially modifiable dementia risk factors (6), based on growing evidence (49,50).

Hearing loss was assessed based on the participant's own perception of their hearing, including if they used a hearing aid. It was self-reported using the question: "How would you rate your hearing (even when using a hearing aid)?" Response options included: very good or excellent, good, fair, poor, and very poor. Following the approach of a previous study using ELSI-Brazil data, responses were categorized into: good (including "very good or excellent" and "good"), fair, and poor (including "poor" and "very poor") (51).

### 2.4.4.3 Physical Activity Level, Handgrip Strength, and Lifestyle Habits

To assess the level of physical activity, the short version of the *International Physical Activity Questionnaire* (IPAQ) was used (52), translated and validated for Brazil (53,54) and for use among adults and older individuals (55). This questionnaire includes questions regarding frequency (days) over the past week and duration (minutes) of activities such as walking (low intensity), moderate, and/or vigorous-intensity exercise. The *Metabolic Equivalent of Task* (MET) was used as a measure to weight each type of activity according to its energy cost, expressed as a multiple of the resting metabolic rate. A *Total Physical Activity Score* (TPA) in MET-minutes/week was calculated by multiplying the MET score of each activity (3.3 for walking, 4.0 for moderate, and 8.0 for vigorous activity) by the minutes performed and weekly frequency. MET-minutes/week were calculated separately for each activity and then summed. A categorical score was then created based on the number of days, duration, type of activity, and TPA. Cases in which responses included "Don't know/Did not answer" were treated as missing data and excluded from analysis. All steps followed the *IPAQ Data Processing and Analysis Guidelines* for categorizing physical activity levels (56).

Handgrip Strength (HGS) was measured using a hydraulic hand dynamometer with an adjustable handle (SAEHAN®, South Korea, JAMAR). The test was conducted three times with a one-minute rest interval between each measurement. Participants chose their dominant hand for the test, which was performed while seated in an armless chair, with the elbow flexed at 90 degrees, the arm alongside the body without support or additional movement during the test, the forearm in a neutral position (thumb pointing up), and the wrist in a comfortable position. The dynamometer handle was placed in the second position or adjusted according to hand size when necessary. During the test, participants were given verbal encouragement as sensory stimuli (57,58).

Cut-off points for HGS categorization were established according to international norms (59), adjusted by sex and age using percentile distributions of absolute strength (kgf) reported in that study. For individuals aged 50 and older, handgrip strength was classified into five categories: *low* (below the 20th percentile), *slightly low* (20th–39th percentile), *moderate* (40th–59th percentile), *slightly high* (60th–79th percentile), and *high* (80th percentile or higher). Absolute values (in kgf) for men and women are presented in Supplementary Table 4. Responses such as "Refused" or "Did not attempt due to perceived risk" were treated as missing data. For responses like "Attempted but couldn't" or "Unable," a score of 0 was assigned, as the HGS test is a functional measure and inability to perform the test reflects compromised health and a complete absence of strength. Furthermore, including HGS in dementia prediction models can improve accuracy by incorporating an objective indicator of physical health that reflects both central nervous system integrity and musculoskeletal condition.

Smoking was assessed through self-report using the question: "Do you currently smoke?" The question was presented within a broader context to capture the widest range of smoking behaviors, including manufactured cigarettes, straw cigarettes, or other smoked tobacco products (cigars, cigarillos, pipes, clove or Bali cigarettes, Indian cigarettes or bidis, and hookahs/water pipes). Tobacco products such as snuff, chewing tobacco, and electronic cigarettes were not considered. Response options were: "Yes, daily," "Yes, less than daily," "No," and "Don't know/Did not answer." The last option was treated as missing data.

Alcohol consumption was also self-reported based on the following questions: "How many days per week do you usually drink any alcoholic beverage?" and "On a typical day when you drink, how many drinks do you consume?" For the first question, responses ranged from 1 to 7 days, with 0 for individuals who drank less than once a week. The second question referred to the number of drinks per day, with 1 drink equivalent to 1 can of beer, 1 glass of wine, or 1 shot of cachaça, whiskey, or other distilled beverages. Individuals who responded "Don't know/Did not answer" to either question were excluded as missing data. Excessive alcohol consumption was defined by calculating the total number of drinks per week (drinks per day × days per week), with 21 or more drinks per week (equivalent to 168 grams of pure alcohol) considered excessive (6,39).

### 2.4.4.4 Psychosocial Factors

Social contact was assessed through questions on social relationships. Three items were used: "How often do you meet in person with any of your children, not counting those who live with you?", "How often do you meet in person with any of your relatives, not counting those who live with you?", and "How often do you meet in person with any of your friends, not counting those who live with you?" Possible responses included: "3 or more times per week," "1 or 2 times per week," "1 or 2 times per month," "Every 2 or 3 months," "Once or twice per year," and "Less than once a year or never."

To identify low social contact/social isolation, an individual was considered socially isolated if they reported meeting with children, relatives, or friends less than once a month (39). Otherwise, participants were categorized as having social contact. If the participant answered "Don't know/Did not answer" to all three questions, their data were treated as missing and excluded from the social isolation analysis. However, if they responded to at least one of the three items, they were included, since social contact frequency was defined based on the available response(s).

Loneliness was self-reported based on the question: "How often do you feel lonely?" Possible responses were: "Never," "Sometimes," and "Always." Self-reported loneliness was considered present when the participant indicated insufficient social contact according to their own subjective experience. Responses of "Don't know/Did not answer" were treated as missing data.

Subjective well-being (SWB) was evaluated using a life satisfaction ladder. Participants were shown an illustration of a ten-rung ladder representing general life satisfaction, ranging from 1 (lowest level) to 10 (maximum satisfaction). Participants were asked to reflect on their overall SWB and indicate the rung that best represented their current satisfaction. The interviewer recorded the number selected.

Depressive symptoms were assessed using the eight-item version of the Center for Epidemiologic Studies Depression Scale (CES-D8) (28,60), a validated short form (32,61,62). The CES-D8 includes the following items: "During the past week, most of the time, did you feel depressed?", "During the past week, most of the time, did you feel that everything you did was more difficult than usual?", "During the past week, most of the time, did you feel that your sleep was not restful, that did you wake up not feeling rested?", "During the past week, most of the time, did you feel happy?", "During the past week, most of the time, did you feel lonely?", "During the past week, most of the time, did you enjoy life or feel pleasure from it?", "During the past week, most of the time, did you feel sad?" e "During the past week, most of the time, did you feel that you couldn't get going with your activities?". Each item could be answered with "Yes" or "No." For items 1, 2, 3, 4, 6, and 8, a "Yes" response scored 1 point. For items 5 and 7, a "No" response scored 1 point. The total score ranges from 0 to 8. A score of 4 or higher was used as the cutoff to indicate the presence of depressive symptoms, while scores below 4 indicated absence of symptoms. Participants who answered "Don't know/Did not answer" or "Not applicable" were excluded from the analysis of depressive symptoms, and these data were treated as missing.

## 2.5 Statistical Analysis

The analyses were conducted by the researchers using the Python programming language, version 3.10, through the online platform Google Colaboratory (63). The following libraries were used: Numpy version 1.26.4, Pandas version 2.2.2, seaborn version 0.13.2, statsmodels version 0.14.4, scipy version 1.13.1, matplotlib version 3.10.0, and scikit-learn version 1.7.1 (64–67).

Continuous variables that were not normally distributed (as assessed by the Shapiro–Wilk test) were reported as medians and interquartile ranges; comparisons were performed using the Mann–Whitney U test. Categorical variables were reported as counts and percentages and compared using the chi-square test. For all statistical tests, a p-value < 0.05 was considered statistically significant (68).

The application of the Random Forest (RF) algorithm enabled the analysis of large volumes of complex data, identifying patterns and combinations between variables that might be overlooked by traditional methods, in addition to its ability to handle non-linear data (69). Although other feature selection techniques exist, RF was chosen for its robustness and versatility in handling various types of data. It is particularly effective for binary, imbalanced datasets with multiple predictors or complex structures, including correlated variables. Furthermore, a key advantage of this algorithm is its intrinsic ability to measure feature importance and use out-of-bag (OOB) error, which provides a reliable estimate of model performance and supports the selection of the optimal number of variables for model construction (70–73). The most important features – those with the lowest average error rates selected by the RF algorithm – were subsequently analyzed using multivariable logistic regression.

### 2.5.1 Data Preprocessing

Initially, before proceeding with any steps in this stage, the dataset was split into training (80%) and testing (20%) sets, while preserving the class proportions. All subsequent steps were applied exclusively to the training set, with the test set reserved for later model performance evaluation.

In the training set, missing data were identified for each variable. The K-Nearest Neighbors (KNN) multiple imputation method (74) was used to replace each missing value with an observed value from similar records in the dataset, identified using a distance metric. In this case, we used the default KNN metric: Euclidean distance, which calculates the square root of the sum of squared differences between non-missing values, assessing similarity between individuals. The closest neighbors were then used to fill in the missing values with the average of their corresponding values. This is crucial in data analysis, as it helps preserve the original data structure and prevents distortion in the distribution of imputed variables (74,75).

This algorithm is widely applied in health research to address missing data, which can compromise analysis quality, ensuring data integrity and representativeness (77). In the medical field, KNN imputation improves the accuracy of predictive models and supports decision-making, for example, in heart disease datasets, where this method has shown improved performance, provided a proper balance is maintained between the number of neighbors used and the preservation of the original data structure (77,78).

Before handling missing values, data normalization was performed. At this point, only the *life satisfaction* variable was normalized using MinMaxScaler, as it is a continuous variable (ranging from 0 to 10). In the KNN algorithm, such variables can disproportionately influence the distance metric and, consequently, affect the selection of nearest neighbors and the imputation of all other variables.

Next, the Elbow Method (79,80) was applied to determine the optimal number of neighbors ($k$) for KNN imputation. This technique involves testing different values of $k$ and analyzing a graph to identify the point at which increasing $k$ no longer results in meaningful performance improvement. This point, known as the "elbow," corresponds to the ideal value for $k$ based on the curve's shape.

This method was tested on a dataset in which 10% of the values were randomly removed and replaced with NaN (Not a Number), simulating the presence of missing data. This allowed for an evaluation of the imputation method and the determination of the $k$ value that minimizes the Mean Absolute Error (MAE).

The training set (with no missing values) was divided into two subsets to enable comparison between actual and imputed values in the simulated dataset, allowing for MAE calculation. Values of $k$ from 1 to 20 were tested to explore different configurations. For each $k$, a KNN imputer was created, and missing values were imputed. MAE was then computed by comparing imputed values with their actual counterparts.

This entire process was repeated 30 times to increase result robustness. The errors from each iteration were stored in a list, enabling the calculation of the average error for each $k$ value. The point at which the error reduction curve shows a significant slowdown, the "elbow", was selected as the optimal number of neighbors. This approach balances imputation accuracy with model simplicity, allowing for efficient estimation of missing data without unnecessary complexity (81).

After determining the optimal *k*, the KNNImputer algorithm (82) was employed to fill in the missing values based on similar individuals in the dataset (nearest neighbors). This algorithm is part of the scikit-learn library (64), a widely used Python library containing pre-built tools for data science and machine learning tasks. In this context, a library is a digital repository of reusable algorithms that facilitate data analysis. The module sklearn.impute (83) within scikit-learn specializes in methods for handling missing data (imputation techniques).

### 2.5.2 Model Building

RF (84) combines multiple classifiers through parallel decision trees, where each tree is built using a random vector sampled independently and identically distributed for all trees in the forest. At each split node in a tree, a random subset of features is selected to determine the best split. Each tree in the forest produces a classification, and the final decision is made by aggregating the outputs of all trees through majority voting. This method is known as bagging or bootstrap aggregating.

Feature importance analysis was based on the reduction in impurity during the construction of decision trees. This process is measured using the Gini index, which quantifies how homogeneous the data are with respect to the target variable (cognitive status). Each time a variable is used to split a node, the decrease in the Gini index is calculated, and the overall importance of a variable is obtained by summing all such reductions across all trees in the forest. The greater this total reduction, the more important the variable is to the model, providing a direct and interpretable estimate of the variable's relevance to model decisions.

After ranking the variables by importance, we aimed to identify the smallest subset of variables that minimized the out-of-bag (OOB) error, an intrinsic metric of the RF algorithm. OOB error is computed during training using data not selected for building each tree – this forms the OOB sample. For each individual in the OOB sample, predictions are made only using trees that did not have access to that individual's data (by aggregating the votes from these trees). OOB accuracy is then calculated as the proportion of correct predictions, and OOB error is defined as 1 minus OOB accuracy.

In this process, variables were added incrementally in order of importance to perform a stepwise RF analysis. Model training began with only the most important variable, followed by the addition of the second most important, and so on, until all variables (in order of decreasing importance) were included. For each subset of variables, training was repeated multiple times using a stratified 10-fold cross-validation approach (85) with 30 repetitions, accounting for 300 experiments per iteration.

This technique involves splitting the dataset into multiple subsets, known as folds, allowing the model to be trained repeatedly on different portions of the training set. This approach is essential for estimating the model's true performance on unseen data, minimizing both bias and variance during training, and ensuring a more robust evaluation.

However, instead of using a separate validation set, we relied on the out-of-bag (OOB) error metric from the RF as the validation method, calculating error on OOB samples. Using this approach, for each additional variable included in the model, we obtained the mean and variability of the OOB error. At the end of this process, we were able to identify the subset of variables that minimized the mean OOB error, thereby determining the most parsimonious and best-performing set of predictors (86).

To build the final predictive model estimating the likelihood of developing dementia, the predictors with the highest importance scores and lowest average OOB error rates were analyzed using multivariable logistic regression, allowing for mutual adjustment between variables and yielding coefficients that were converted into odds ratios (ORs) to enhance model interpretability. Statistical significance was considered at $p < 0.05$. The full model development process is illustrated in Figure 1.

Figure 1. Process used for model construction.

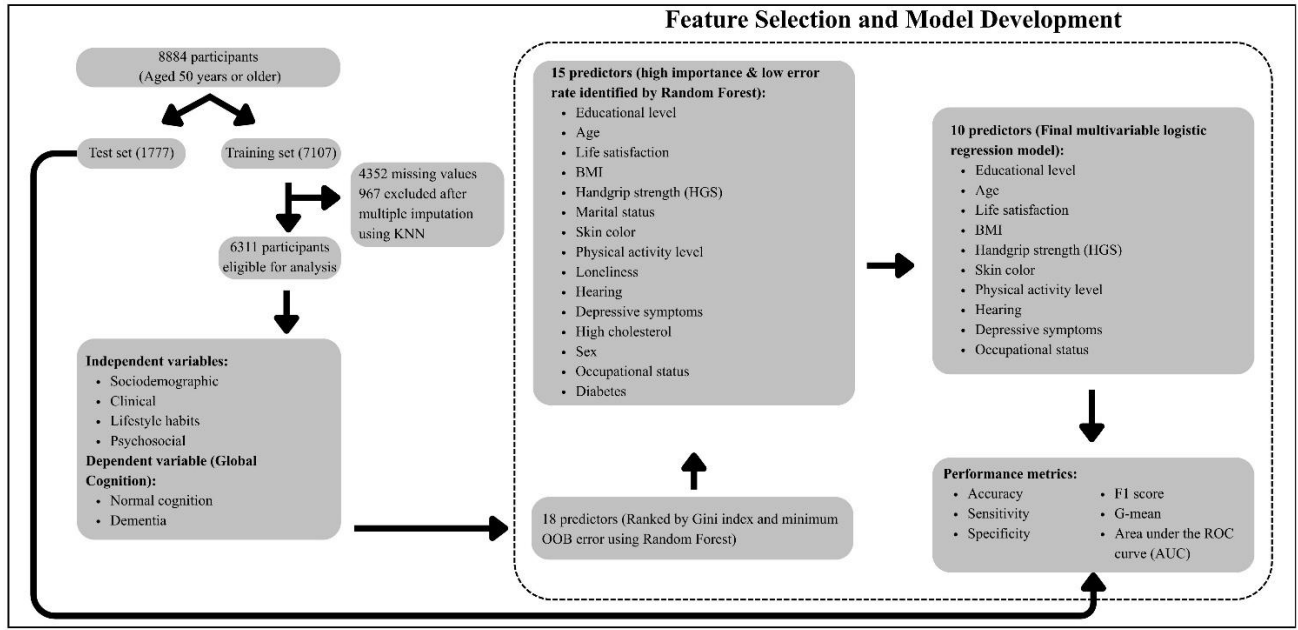

After performing multiple logistic regression analysis, we identified the predictors that were significantly associated with dementia. Using this subset of variables, we retrained the models from scratch on the entire training dataset, employing both the Random Forest (RF) algorithm and logistic regression for comparison purposes. Finally, we evaluated the performance of both models on the test dataset.

## 3 Results and Discussion

### 3.1 Results

#### 3.1.1 General Information

Out of all 8,884 participants, 845 had dementia (9.6%). Except for hypertension, cataract, and excessive alcohol consumption (which were excluded), there were statistically significant differences across all potential predictors (see Supplementary Table 5).

#### 3.1.2 Response Variable Categorization

The multiple regression analysis revealed that age and education were significant predictors of the Standardized Global Cognitive Score (SGCS) in the analyzed subsample (n = 1,185). The model explained 34.6% of the variance in SGCS ($R^2 = 0.346$). While not exceptionally high, this indicates that education and age have a substantial impact on cognitive performance. Furthermore, the model demonstrated statistically significant explanatory power (F-statistic = 49.16, p < 0.000). Specifically, age was inversely correlated with SGCS (coefficient = -0.268, p < 0.000), whereas higher education levels were directly correlated (coefficient = 0.175, p < 0.000). No significant differences were found between sexes (coefficient = 0.044, p = 0.128).

### 3.1.3 Data Preprocessing

Missing values were present in 13 predictors (see Supplementary Material 1). The proportion of missing data ranged from 10.8% to 0.1%. Since this could affect the quality of the analyses and the reliability of model results, we adopted a multiple imputation approach using the KNNImputer algorithm, beginning with the Elbow Method to determine the optimal number of neighbors. The Mean Absolute Error (MAE) ranged from 0.035 to 0.031, reflecting good overall performance of the imputation algorithm and aiding in the selection of the optimal *k* value (see Supplementary Material 1). In addition, the data were complete for the predictors sex/gender, age, educational level, social isolation, and occupational status.

### 3.1.4 Random Forest Analysis

We analyzed the importance ranking of 18 predictors using the RF algorithm, covering sociodemographic, clinical, lifestyle, and psychosocial data, with dementia as the dependent variable. The results showed that the top five predictors, in descending order of mean decrease in impurity, were: educational level, age, life satisfaction, BMI, and HGS. The complete results can be seen in Figure 2.

Figure 2. Importance ranking of variables influencing dementia in individuals aged 50 years and older.

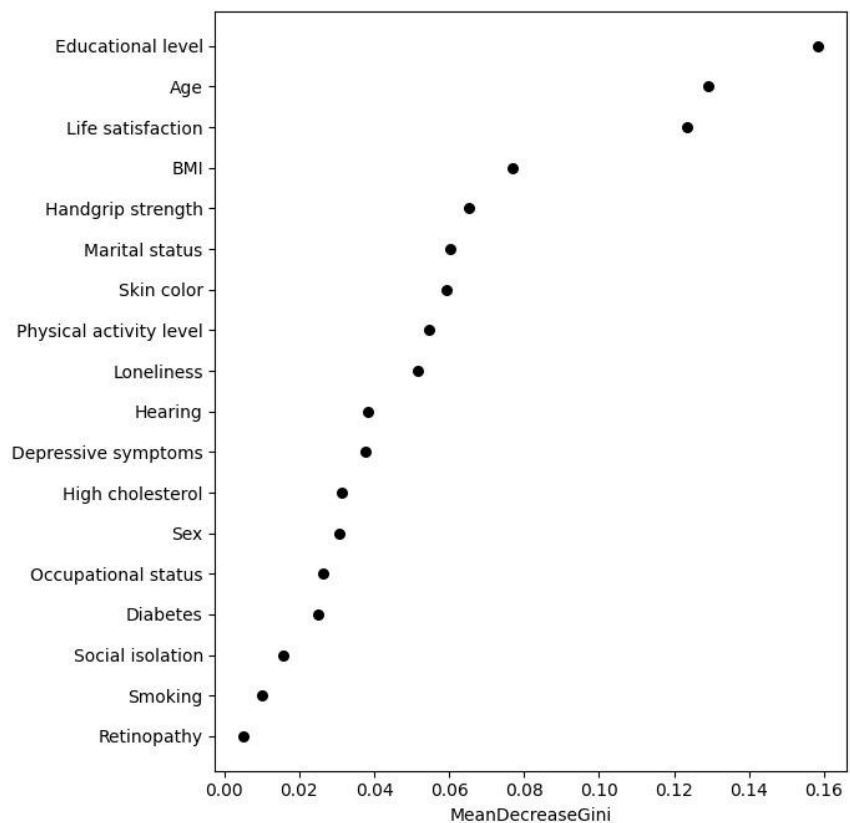

Based on the results of the predictor importance ranking, we analyzed the mean out-of-bag (OOB) error rate as predictors were gradually added, in order to identify the subset that minimized error and enabled dimensionality reduction. As shown in Figure 3, the lowest OOB error rate of 0.0812% (± 0.0006) was achieved. All values are presented in Supplementary Table 6. This indicates that the first 16 predictors, ranked by importance, yielded a low error rate in the data analysis. These predictors were: educational level, age, life satisfaction, BMI, HGS, marital status, skin color, physical activity level, loneliness, hearing, depressive symptoms, high cholesterol, sex, current occupational status, and diabetes.

Figure 3. Out-of-Bag Error Rate.

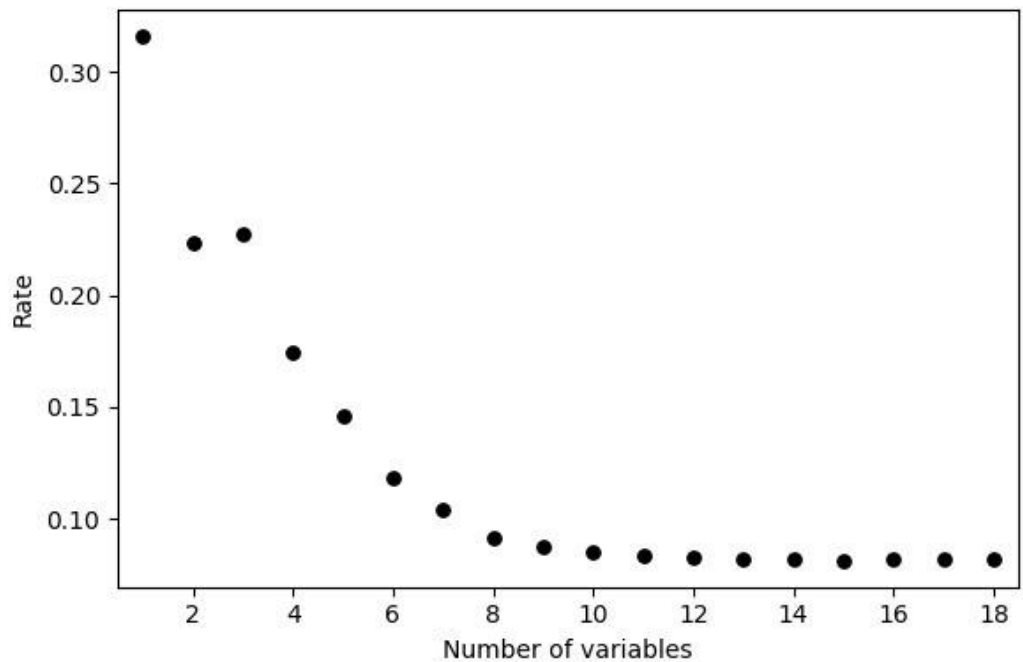

### 3.1.5 Multivariable Logistic Regression Analysis

After the initial analysis using RF, the 16 most important independent variables were selected for a multivariable logistic regression analysis, with the goal of developing an interpretable model, using dementia as the dependent variable.

The variables and their respective values are presented in Supplementary Table 7. The analysis showed statistically significant results for educational level, age, life satisfaction, BMI, HGS, skin color, physical activity level, hearing, depressive symptoms, and current occupational status, all of which influenced dementia in adults aged 50 years and older (see Table 1).

### 3.1.6 Model Performance

Using the test set, final models were evaluated based on the 10 independent predictors that were significantly associated with dementia. The Random Forest (RF) model achieved an Area Under the ROC Curve (AUC-ROC) of 0.776 (95% CI: 0.740–0.811), with sensitivity of 0.708, specificity of 0.702, F1 score of 0.311, G-means of 0.705, and accuracy of 0.703. The logistic regression model showed a similar but slightly lower performance, with an AUC-ROC of 0.735 (95% CI: 0.697–0.773), sensitivity of 0.678, specificity of 0.654, F1 score of 0.272, G-means of 0.666, and accuracy of 0.656.

Table 1. Multivariable Logistic Regression Model Analysis

| Variable | Normal Cognition (N=7962) | Dementia (N=845) | OR (95% CI) | P |
|---|---|---|---|---|
| Educational level, n (%) | | | | |
| Illiterate | 1031 (12.9) | 332 (39.2) | 7.42 (4.04-13.62) | <0.001 |
| Less than Primary Education | 1548 (19.4) | 217 (25.6) | 3.42 (1.88-6.22) | <0.001 |
| Completed Primary Education | 2472 (31.0) | 199 (23.5) | 1.90 (1.05-3.43) | 0.0343 |
| Incomplete Secondary Education | 936 (11.7) | 46 (5.4) | 1.16 (0.59-2.27) | 0.6609 |
| Completed Secondary Education | 1395 (17.5) | 28 (3.3) | 0.44 (0.21-0.94) | 0.0337 |
| Higher Education or More | 580 (7.2) | 23 (2.7) | - | - |
| Age, n (%) | | | | |
| 50-54 | 1927 (24.2) | 134 (15.8) | - | - |
| 55-59 | 1499 (18.8) | 166 (19.6) | 1.62 (1.19-2.20) | 0.0019 |
| 60-64 | 1302 (16.3) | 155 (18.3) | 1.42 (1.04-1.96) | 0.0298 |
| 65-69 | 1028 (12.9) | 144 (17.0) | 1.27 (0.90-1.79) | 0.1764 |
| 70-74 | 878 (11.0) | 80 (9.4) | 0.65 (0.43-1.02) | 0.0406 |
| 75-79 | 712 (8.9) | 48 (5.6) | 0.28 (0.16-0.48) | <0.001 |
| 80-84 | 393 (4.9) | 28 (3.3) | 0.15 (0.06-0.37) | <0.001 |
| 85-89 | 187 (2.3) | 27 (3.2) | 0.28 (0.10-0.76) | 0.0125 |
| 90+ | 36 (0.4) | 63 (7.4) | 11.01 (5.06-23.95) | <0.001 |
| Life satisfaction, median (IQR) | 8.0(5.0-10.0) | 7.0 (4.0-10.0) | 0.72 (0.52-0.99) | 0.0462 |
| BMI, n (%) | | | | |
| Severely underweight | 31 (0.4) | 4 (0.5) | 0.14 (0.01-1.79) | 0.1316 |
| Underweight | 109 (1.4) | 26 (3.4) | 2.11 (1.12-3.97) | 0.0210 |
| Normal weight | 2110 (27.5) | 271 (35.8) | - | - |
| Overweight | 3097 (40.3) | 257 (33.9) | 0.82 (0.64-1.03) | 0.0912 |
| Moderately obese (Grade I) | 1628 (21.2) | 142 (18.7) | 0.78 (0.58-1.05) | 0.1011 |
| Severely obese (Grade II) | 502 (6.5) | 42 (5.5) | 0.76 (0.49-1.18) | 0.2273 |
| Morbid obesity (Grade III) | 191 (2.4) | 14 (1.8) | 0.67 (0.32-1.38) | 0.2736 |
| Handgrip strength, n (%) | | | | |
| High | 254 (3.3) | 11 (1.3) | - | - |
| Slightly high | 823 (10.7) | 46 (5.7) | 1.68 (0.68-4.11) | 0.2591 |
| Moderate | 1234 (16.0) | 83 (10.4) | 1.62 (0.68-3.85) | 0.2750 |
| Slightly low | 2217 (28.8) | 171 (21.4) | 1.74 (0.75-4.03) | 0.1995 |
| Low | 3162 (41.1) | 485 (60.9) | 2.50 (1.09-5.76) | 0.0313 |
| Marital status, n (%) | | | | |
| Single | 848 (10.6) | 112 (13.2) | 1.00 (0.72-1.39) | 0.9771 |
| Married/partner/stable union | 4705 (59.0) | 410 (48.5) | - | - |
| Divorced or separated | 978 (12.2) | 98 (11.6) | 0.77 (0.56-1.06) | 0.1147 |
| Widowed | 1431 (17.9) | 225 (26.6) | 1.11 (0.83-1.48) | 0.5003 |
| Skin color, n (%) | | | | |
| White | 3157 (40.9) | 248 (31.3) | - | - |
| Black | 713 (9.2) | 114 (14.4) | 1.47 (1.07-2.00) | 0.0157 |
| Brown (mixed race) | 3589 (46.5) | 395 (50.0) | 1.04 (0.83-1.31) | 0.7306 |
| Yellow (East Asian origin) | 73 (0.9) | 9 (1.1) | 1.26 (0.51-3.09) | 0.6198 |
| Indigenous | 183 (2.3) | 24 (3.0) | 1.01 (0.52-1.97) | 0.9660 |
| Physical activity level, n (%) | | | | |
| High | 2330 (30.9) | 149 (19.3) | - | - |
| Moderate | 2190 (29.1) | 155 (20.1) | 1.14 (0.87-1.50) | 0.3501 |
| Low | 3004 (39.9) | 467 (60.5) | 1.61 (1.25-2.08) | <0.001 |
| Loneliness, n (%) | | | | |
| Never | 3765 (52.0) | 276 (42.8) | - | - |
| Sometimes | 2349 (32.4) | 165 (25.6) | 0.93 (0.72-1.19) | 0.5538 |
| Always | 1124 (15.5) | 203 (31.5) | 1.23 (0.93-1.64) | 0.1529 |
| Hearing, n (%) | | | | |
| Good | 5554 (69.8) | 525 (62.4) | - | - |
| Fair | 1985 (24.9) | 197 (23.4) | 0.93 (0.74-1.18) | 0.5675 |
| Poor | 416 (5.2) | 119 (14.1) | 1.65 (1.16-2.37) | 0.0059 |
| Depressive symptoms, n (%) | | | | |
| No | 4904 (66.8) | 302 (45.9) | - | - |
| Yes | 2434 (33.1) | 355 (54.0) | 1.72 (1.36-2.16) | <0.001 |
| Sex, n (%) | | | | |
| Male | 3522 (44.2) | 312 (37.0) | - | - |
| Female | 4440 (55.7) | 532 (62.9) | 1.07 (0.86-1.34) | 0.5531 |
| High cholesterol, n (%) | | | | |
| No | 5453 (69.0) | 554 (66.6) | - | - |
| Yes | 2447 (30.9) | 277 (33.3) | 1.12 (0.91-1.39) | 0.2805 |
| Occupational status, n(%) | | | | |
| No (Not working) | 5465 (68.6) | 691 (81.7) | - | - |
| Yes (working) | 2497 (31.3) | 154 (18.2) | 0.78 (0.61-1.00) | 0.0463 |
| Diabetes, n (%) | | | | |
| No | 6669 (84.1) | 666 (79.5) | - | - |
| Yes | 1259 (15.8) | 171 (20.4) | 1.02 (0.78-1.33) | 0.8986 |

Legend: BMI: Body Mass Index; IQR: Interquartile Range.

## 3.2 Discussion

In this study, we used data collected from the Brazilian Longitudinal Study of Aging (ELSI-Brazil), one of the largest population-based studies in Brazil, with national representation of non-institutionalized individuals aged 50 years and older. Our objective was to develop a simple predictive model to estimate the likelihood of developing dementia in middle-aged and older Brazilian adults, by analyzing sociodemographic, health, lifestyle, and psychosocial factors as predictors and identifying which of them were most relevant in building the model.

General characteristics of the sample are presented in Supplementary Table 5. Initially, the study included 9,412 participants. After applying the neuropsychological assessments and informant-based cognitive function evaluation, we were able to classify individuals into three cognitive status groups: normal cognition, cognitive impairment, and dementia. For the purposes of this analysis, we included only participants classified as having normal cognition or dementia, resulting in a total of 8,807 participants with complete data.

### 3.2.1 Prevalence of Dementia Among Middle-Aged and Older Adults in Brazil

The results of this study revealed that the prevalence of dementia among Brazilian adults aged 50 and older was 9.6%, a value higher than that reported in a recently published study, which estimated the prevalence of all-cause dementia in Brazil at 8.5% (87). However, it is important to note that the authors of that study included only individuals aged 60 and over, whereas the present study assessed cognition beginning at age 50. Although dementia traditionally affects older individuals, there is growing evidence that it may manifest – albeit in milder forms – at increasingly earlier ages (88–90).

A general trend of increasing dementia prevalence with advancing age was observed. The lowest prevalence occurred among participants aged 50 to 54 years (6.5%), while the highest was found in the 90+ age group, reaching 62.4%. This finding is consistent with well-established knowledge that dementia predominantly affects individuals in older age groups (91).

As in the study by Alibert et al. (87), which reported higher prevalence among women (9.1%) compared to men (7.7%), our results also reflect this trend, with rates of 10.7% for women and 8.2% for men. These findings align with multiple studies suggesting that women face an increased lifetime risk of developing dementia, potentially due to biological, social, and cognitive factors (92–95).

We also analyzed dementia prevalence by educational level and found that it was highest among illiterate participants (24.4%), approximately two to three times higher than in intermediate education levels with 12.3% for those with less than primary education and 7.5% for those who completed primary school. Prevalence was 4.7% among those with incomplete secondary education and lowest among those with completed secondary education (2.0%). A slight increase was observed in the group with higher education or more (3.8%). This inverse pattern – lower prevalence with higher educational attainment – reinforces findings from previous studies in Brazil and elsewhere, in which education is shown to serve as a marker of cognitive reserve and a protective factor against dementia onset (96,97).

### 3.2.2 Random Forest Algorithm and Multivariable Logistic Regression Analysis of Dementia Among Middle-Aged and Older Adults in Brazil

The RF algorithm and a multivariable logistic regression analysis were conducted in this study to investigate factors influencing dementia in Brazilian adults aged 50 years and older. After adjustments, the results revealed that the odds of developing dementia were 7.41 times higher (OR = 7.41; 95% CI: 4.03–13.62; $p < 0.001$) for illiterate individuals. As educational level increased, the odds decreased, though they remained elevated for those with low education. Individuals with less than primary education had 3.41 times greater odds (OR = 3.41; 95% CI: 1.87–6.21; $p < 0.001$), and those who completed primary education had 89% higher odds (OR = 1.89; 95% CI: 1.04–3.42; $p = 0.034$). In contrast, those who completed high school showed a 57% reduction in the odds (OR = 0.43; 95% CI: 0.20–0.93; $p = 0.033$) compared to individuals with a university degree or higher. The group with incomplete high school, although showing lower odds (OR = 1.16), did not exhibit a statistically significant association (95% CI: 0.59–2.27; $p = 0.660$). Although no consistent evidence suggests structural brain differences between individuals with high versus low education levels (98), the influence of lifelong intellectual stimulation and cognitive complexity is well established. This supports the hypothesis that higher education levels are associated with a reduced risk of dementia (98), possibly due to mechanisms related to cognitive reserve (99,100). However, the findings in this study may reflect not only the absence of formal education – as typically categorized in epidemiological studies – but also broader structural inequalities, such as poverty, limited access to healthcare, less cognitively demanding occupations, and reduced engagement in cognitively stimulating activities throughout life (100). Thus, the greater likelihood of dementia among individuals with low education may not solely reflect the number of years spent in school, but rather a lifelong trajectory marked by cumulative disadvantage, which affects both cognition and brain health.

The relationship between age and dementia is often complex, influenced by various interacting factors. However, dementia prevalence generally increases with advancing age across all causes (101). In our findings, individuals aged 90 and above exhibited significantly higher odds of dementia compared to the youngest age group (OR = 11.00; 95% CI: 5.05–23.95; $p < 0.001$), which aligns with well-established literature (101). Interestingly, this trend was not observed in the age groups between 70 and 89 years. In fact, the odds ratios indicated lower chances of dementia in individuals aged 70–74 years (OR = 0.65; 95% CI: 0.43–1.02; $p = 0.0406$), although the confidence interval includes an OR of 1.00 (suggesting equal odds between groups). The odds were also significantly lower for those aged 75–79 years (OR = 0.28; 95% CI: 0.16–0.48; $p < 0.001$), 80–84 years (OR = 0.15; 95% CI: 0.06–0.37; $p < 0.001$), and 85–89 years (OR = 0.28; 95% CI: 0.10–0.76; $p = 0.0125$). This seemingly paradoxical association can be explained through the lens of survivor bias. According to mortality data from DATASUS for the years 2015–2016, all-cause mortality increased from 58.5 per 100,000 inhabitants in the 70–74 age group to 182.5 in individuals aged 80 and over (102). Survivor bias arises when individuals who live to older ages and remain cognitively intact represent a more resilient subgroup due to favorable biological, behavioral, or social characteristics. Conversely, more vulnerable individuals may have developed dementia-related pathologies earlier and died before reaching older ages, thus being excluded from the observed sample. This results in an artificial underrepresentation of dementia cases in older age brackets, thereby skewing the distribution and underestimating the true odds of dementia (103). Statistically, high mortality functions as a selective filter – reducing the denominator in logistic regression models – which can lead to artificially lower odds ratios, creating the illusion of a protective effect of advanced age. Although this interpretation may be plausible, it cannot be confirmed, given that the data represents only a cross-sectional snapshot in time. Therefore, this explanation should be understood as an interpretative and speculative hypothesis. Nonetheless, the significant association observed in individuals aged 90 and over reinforces the fact that severe vascular and neurological changes associated with aging continue to play a critical role in the development of dementia at very advanced ages (104). It is important to note that despite the apparent protective pattern in intermediate age groups, these results should be interpreted with caution. The cross-sectional nature of this study limits causal inference and introduces the potential for bias. Longitudinal studies are needed to better elucidate the trajectory of age-related cognitive decline and to support the implementation of preventive strategies earlier in life.

Our findings indicate that engaging in any occupational activity, such as paid work in the past 30 days, was associated with a 22% reduction in the odds of dementia compared to not performing occupational activities (OR = 0.78; 95% CI: 0.61–1.00; p = 0.0463). This suggests a potential protective effect of occupational activity, with the best-case scenario (based on the lower confidence limit) indicating up to 39% lower odds. However, the confidence interval shows that in the worst-case scenario, the clinical relevance may be negligible, as the upper limit of the interval reaches OR = 1.00. Additionally, the result lies at the threshold of statistical significance, indicating a weak or potentially null association. Recent evidence assessing employment trajectories between the ages of 18 and 65 indicates that not working, whether due to disability or unemployment, is associated with poorer cognitive performance in later life (105). This may occur because cognitive stimulation and sustained productive engagement, beyond formal education, reduce brain vulnerability through cognitive protection mechanisms. Moreover, work promotes autonomy and fosters long-term social exposure, which directly impacts social networks (105,106). Although our findings suggest only a weak association, they support the hypothesis of functional cognitive reserve. Performing cognitively demanding tasks or maintaining occupational routines may help delay or reduce the manifestation of cognitive decline (107,108). It is important to note that our indicator captures only whether an individual is currently engaged in occupational activity, it does not assess the type of job, the complexity of tasks, or the duration of occupational exposure, which are key factors that influence cognitive demands. These factors may also affect levels of proteins that inhibit neuroprotective mechanisms, altering brain structure through synaptic plasticity and functional connectivity (i.e., synaptogenesis and axonogenesis) (106,108). Nonetheless, these findings underscore the importance of public policies aimed at promoting or maintaining productive and occupational activities throughout the life course, as a critical strategy for supporting brain and cognitive health.

In addition to maintaining occupational engagement throughout life, another key social marker of vulnerability, skin color, emerges as a relevant factor in our analysis, revealing a structural inequality that directly impacts brain and cognitive health. Our findings indicate that self-identified Black individuals had significantly higher odds of dementia compared to self-identified White individuals (OR = 1.47; 95% CI: 1.07–2.00; p = 0.0157). This result aligns with previous Brazilian studies highlighting ethnic-racial disparities in dementia burden, including higher dementia-related mortality rates among Black populations (109). Interestingly, cognitive and functional abilities appear to be similar between Black and White individuals, and race has not been shown to modify the associations between cognition and neuropathology (110). Therefore, our results are likely to race as a marker of socio-structural inequality rather than a biological determinant. In Brazil, Black populations continue to face perceived barriers in accessing healthcare services (111,112), lower educational attainment with higher dropout rates in basic education, and limited access to more qualified employment opportunities, even when educational credentials are comparable to those of White individuals (113). These findings highlight the presence of social and structural vulnerabilities, notably low educational attainment, economic instability, and social insecurity, all of which are social determinants that negatively affect healthy brain aging in the Brazilian population (114).

Aging is characterized by a series of changes resulting from cumulative psychosocial, physiological, and neurological processes throughout life, which can negatively impact cognitive health. Among these changes, sensory loss, such as reduced hearing acuity, has gained increasing attention as one of the most relevant modifiable risk factors for dementia prevention (7). Self-perceived hearing ability is a critical indicator of cognitive health. In our analysis, participants who reported regular hearing showed no statistically significant association with dementia (OR = 0.93; 95% CI: 0.74–1.18; p = 0.5675). However, individuals who reported poor hearing had a 65% higher likelihood of dementia compared to those who perceived their hearing as good (OR = 1.65; 95% CI: 1.16–2.37; p = 0.0059). Recent evidence from independent samples has confirmed this association (115,116), showing that sensory impairment may accelerate global cognitive decline. Moreover, it is estimated that up to 6.8% of dementia cases could potentially be prevented in the general population if hearing loss were eliminated (117). Several mechanisms may underlie this association, including cognitive overload, in which reduced auditory input consumes neural resources otherwise allocated to memory and attention (118). In addition, prolonged sensory deprivation may induce atrophy in specific brain regions, while shared neural degradation between hearing loss and neurodegeneration may occur via common pathophysiological pathways, such as oxidative stress, mitochondrial dysfunction, and cerebral microangiopathy (119,120). Furthermore, the cognitively impoverished environment caused by hearing loss leads to reduced intellectual and emotional stimulation, contributing to structural changes in the auditory cortex and hippocampus, reduced cognitive reserve, and a direct negative impact on resilience to dementia (118,119). Lastly, translational research has shown multiple alterations in the central nervous system associated with hearing loss, notably reduced neurogenesis in the hippocampus, a brain region critical for memory and auditory processing (121). These findings highlight the importance of early identification of hearing impairment in primary care, aiming to enhance brain health in the Brazilian population.

In this context of modifiable physiological factors that directly influence brain and cognitive health, nutritional status – as reflected by Body Mass Index (BMI) – emerges as an important marker in our analysis. The only statistically significant association was observed among individuals classified as underweight (BMI: 16.0 to 18.4 kg/m²), who had more than twice the odds of dementia compared to those with normal weight (OR = 2.11; 95% CI: 1.12–3.97; p = 0.0210). Although overweight and obesity are currently in the spotlight as major risk factors for dementia (6), recent evidence from various regions, including population-based memory research (PBMR), points to an association between low body weight and dementia, suggesting that underweight status may serve as an early marker of preclinical dementia. However, this association appears to be age-dependent (122,123). Some studies argue that low weight may not be a direct cause of dementia but rather a consequence of it, as weight loss may begin years prior to clinical diagnosis, possibly reflecting metabolic or behavioral changes as subclinical expressions of prodromal dementia (124). Nevertheless, underweight status may also indicate inadequate intake of calories, protein, micronutrients, and minerals, which can compromise several health systems, including the gut microbiome, known to play a key role in cognitive and brain health (125–127). Moreover, underweight individuals are more likely to experience sarcopenia, a condition leading to systemic frailty that affects both physical and cognitive performance (128,129). Therefore, while our findings demonstrate a significant association between low BMI and increased odds of dementia, the nature of this relationship remains uncertain. Further research is needed to clarify this link in adults aged 50 years and older, thereby informing the development of targeted strategies for dementia prevention.

Just as nutrition plays a key role, various lifestyle factors with modifiable potential and direct influence on brain health have gained increasing attention (6), among which physical activity stands out as one of the most consistent pillars in promoting cognitive and brain health (130,131). Our findings demonstrated that individuals with lower levels of physical activity had a 61% higher chance of developing dementia compared to those with high levels of activity (OR = 1.61; 95% CI: 1.25–2.08; p < 0.001). A large body of evidence shows that regular engagement in, or higher levels of, physical activity has protective effects on outcomes related to brain and cognitive health, particularly dementia prevention (132). Not only lower levels of activity, but also a sedentary lifestyle, especially involving passive behaviors for prolonged periods (e.g., watching television), have been associated with reduced physical and cognitive stimulation, thereby increasing the risk of dementia (133). Being physically active plays a crucial role in both physical and mental health. Cunningham et al. (132) identified that physically active older adults had a reduced risk of cognitive decline, dementia, and Alzheimer's disease, as well as lower risks of all-cause mortality, cardiovascular diseases, fractures, disability in activities of daily living, functional limitations, falls, and depression. Beyond its broad health benefits, regular physical activity contributes to a healthier aging trajectory, with better quality of life and cognitive performance (132). Additionally, robust evidence points to exercise-induced changes in brain integrity, including increased total and hippocampal volume driven by neuroplastic mechanisms such as neurotrophic factors, cellular signaling, and growth pathways, directly influencing cognitive functions (e.g., memory and learning), life satisfaction, and psychological well-being (134). Therefore, encouraging and sustaining regular physical activity among middle-aged and older adults in Brazil should be recognized as a public health priority, aimed at strengthening brain health, maintaining cognitive function, and supporting healthy aging.

Within this same framework of physical functionality as a favorable factor for brain health, HGS stands out as an objective marker for assessing muscular strength an element directly associated with functional and cognitive decline in middle-aged and older adults (135). Although the intermediate HGS categories (slightly high, moderate, and slightly low) were not statistically significant in our study, we found a significant association between low HGS (see Supplementary Table 4 for reference values) and an increased risk of dementia. Specifically, individuals with low HGS had 2.5 times higher odds of developing dementia compared to those with high HGS (OR = 2.50; 95% CI: 1.09–5.76; p = 0.0313). The multicenter study coordinated by Zammit et al. (136) revealed a significant correlation between HGS decline and impairments across multiple cognitive domains. This association was further reinforced by findings from Chen (137) and Orchard et al. (138), who demonstrated that individuals with lower HGS had a greater risk of cognitive impairment and dementia. In addition, reduced HGS has been associated with structural changes in the brain, including lower brain volume and atrophy in regions critical to cognition (136). Thus, HGS serves as a low-cost functional marker that reflects the integrity of the central nervous system, brain plasticity, functional reserve, and resilience, all essential elements for healthy brain aging (136). Systematic screening of HGS in middle-aged and older adults, followed by exercise-based interventions aimed at improving muscle strength and reducing functional limitations, may represent a key strategy for the prevention and delay of dementia in the Brazilian population. Such actions could contribute not only to the preservation of functional capacity but also to the promotion of healthy brain aging.

In addition to physical and functional aspects, psychosocial factors also influence brain health, with depressive symptoms and cognitive decline in middle-aged and older adults being consistently investigated worldwide (139,140). Our findings revealed that individuals exhibiting depressive symptoms had a 72% greater likelihood of developing dementia (OR = 1.72; 95% CI: 1.36–2.16; p < 0.001). This result is strongly aligned with a recent meta-analysis, which found that older adults with depression were 1.75 times more likely to develop dementia (141). It is well established that the presence of depressive symptoms significantly interferes with cognitive functioning, including declines in episodic memory, working memory, slower processing speed, and verbal fluency difficulties (142,143). However, the relationship between depressive symptoms and cognitive function appears to be bidirectional: on one hand, changes in depressive symptoms contribute to the decline in cognitive domains, while on the other, deterioration in cognitive performance may intensify depressive symptoms over time (144,145). Both conditions share neurobiological and behavioral mechanisms that affect brain and cognitive health (146), including chronic activation of inflammatory processes – with elevated pro-inflammatory cytokines – and hyperactivity of the hypothalamic-pituitary-adrenal (HPA) axis, resulting in increased cortisol levels and neurotoxic effects, particularly associated with hippocampal volume reduction. Moreover, depression is linked to reduced neurogenesis and synaptic plasticity, impairing cognitive reserve and encouraging behaviors such as physical inactivity, social isolation, and poor adherence to healthy habits, all of which further contribute to cognitive decline. Nevertheless, although depressive symptoms may lead to acute cognitive impairment, this does not necessarily indicate long-term deficits, as improvements in depressive symptoms can attenuate cognitive impairment and, in some cases, even reverse cognitive decline to normal functioning (147–149). Our findings underscore the need for early detection strategies and mental health interventions as essential components of dementia prevention policies. Proper management of depressive symptoms may not only improve quality of life but also represent an effective preventive measure amid the rising global burden of dementia.

Continuing the analysis of psychosocial factors, subjective well-being perception stands out as an important marker of cognitive health. In our findings, life satisfaction emerged as a key psychological indicator, reflecting how content individuals feel with their lives overall or in specific domains. Among middle-aged and older adults, each unit increase in life satisfaction was associated with a 28% lower likelihood of dementia (OR = 0.72; 95% CI: 0.52–0.99; p = 0.0462). This suggests that a positive self-perception of life may serve as a protective factor against cognitive decline. These findings are consistent with longitudinal evidence from various countries, which have demonstrated that greater life satisfaction is associated with a reduced risk of dementia (150,151). This relationship is likely mediated by both psychosocial and behavioral mechanisms. For instance, higher levels of eudaimonic well-being – defined by a sense of purpose, personal growth, self-acceptance, and positive relationships – have been strongly linked to better cognitive functioning and cognitive resilience, the brain's capacity to adapt to stress and recover from mental and emotional adversity (152). Furthermore, family bonds and personal development may play important roles as sources of life satisfaction and perceived life meaning (153). Individuals with higher satisfaction levels are more likely to engage in cognitively stimulating routines, maintain social interactions, and adopt healthy lifestyle habits – factors that contribute to brain health and protect cognition throughout the aging process. In light of these findings, investing in psychosocial interventions that promote subjective well-being – by strengthening social connections, emotional support, and fostering a sense of purpose and self-fulfillment should be considered a strategic element in public health policies aimed not only at improving quality of life but also at supporting healthy aging and potentially reducing the dementia burden in the Brazilian population.

### 3.2.3 Potential Applicability in Primary Health Care

Dementia is one of the leading causes of cognitive impairment and disability among the elderly population and has been steadily increasing due to population aging and modifiable risk factors that contribute to higher prevalence rates (6,39,154). This trend is particularly exacerbated in LMICs (Low- and Middle-Income Countries) such as Brazil, compared to high-income countries (155). The situation in Brazil is especially concerning, with an estimated 80% of dementia cases among individuals aged 60 and older going undiagnosed – particularly in poorer regions and among the younger elderly (ages 60-64 – making timely intervention and management more difficult. This scenario highlights a critical public health challenge (156,157).

Given this context, early detection can significantly improve individual patient management and support more efficient allocation of healthcare resources. Currently, diagnosing dementia is costly and time-consuming, requiring detailed clinical history, physical examination, cognitive assessments, and additional imaging and laboratory tests. With the aging population, the number of new dementia cases is expected to rise, placing even greater strain on healthcare systems. Since a large portion of cases remain undiagnosed, the risk of complications due to inadequate management increases.

Predictive models using machine learning for the detection of cognitive impairment and/or dementia are increasingly being explored worldwide, including growing efforts in LMICs (158,159). However, many of these models rely on high-cost data sources such as imaging, clinical biomarkers, voice analysis, or genetic data, factors that limit their practical implementation in primary care settings (158). An alternative approach, however, involves the use of easily obtainable and low-cost variables, as demonstrated in models developed by Ren et al. (160) using UK data, and Wang et al. (17) using data from China.

Systematic reviews have highlighted a clear gap in the use of clinically accessible variables for dementia prediction (158,161). This underscores the importance of developing models based on practical, easy-to-assess variables that can be collected during routine primary care visits in Brazil using simple exams and standardized questionnaires. A notable example is the AGES-Reykjavik study by Twait et al. (162), which utilized a range of accessible clinical variables to predict dementia in the general population. In their study, the Random Forest algorithm achieved an AUC-ROC of 0.71 (95% CI: 0.68–0.74), with 55% sensitivity and 75% specificity. These results were comparable to those from traditional statistical methods such as logistic regression (AUC-ROC: 0.71; 95% CI: 0.68–0.74), with 64% sensitivity and 68% specificity.

Similarly, our findings showed comparable performance, with the Random Forest model outperforming logistic regression across nearly all metrics. It demonstrated higher discriminative ability (AUC-ROC: 0.776; 95% CI: 0.740–0.811 vs. 0.735; 95% CI: 0.697–0.773), better balance between identifying and excluding cases (sensitivity 70.8% and specificity 70.2% vs. 67.8% and 65.4%), a superior G-mean (0.705 vs. 0.666), and higher overall accuracy (0.703 vs. 0.656). In both models, however, the F1 score remained low (0.311 for Random Forest and 0.272 for logistic regression), indicating modest performance in detecting the positive class and a relatively high rate of false positives.

This study contributes to advancing research in this field by using nationally representative data of Brazilian adults aged 50 and older. It combines the Random Forest algorithm, which handles complex interactions and nonlinearities even in the presence of predictor collinearity, with logistic regression, which allows for clearer interpretation through odds ratio estimates. This hybrid approach enabled the identification of low-cost predictors – sociodemographic, clinical, lifestyle, and psychosocial factors – that are easily assessable in primary care settings and useful for classifying cognitive status in middle-aged and older adults.

The proposed method is not intended for diagnostic purposes but as a rapid and efficient screening tool. It suggests that healthcare professionals could perform an initial risk assessment using brief questions and non-invasive tests.

These findings have important public health implications, particularly for primary care, where early identification of potential neurodegenerative disease can trigger further investigation and health education. Moreover, they support the implementation of early lifestyle interventions, which have shown significant potential in improving cognitive function in individuals with early cognitive impairment or Alzheimer's disease (163), ultimately promoting better quality of life and longevity.

### 3.2.4 Limitations and Implications

These findings enable regular monitoring of potential factors that may contribute to cognitive decline and play a crucial role in encouraging lifestyle changes and, in some cases, early treatment adherence. Such measures can help reduce the burden of cognitive decline and, consequently, dementia in the Brazilian population. By integrating the identification of vulnerable populations and the assessment of these factors into primary healthcare, it is possible to develop effective and accessible prevention strategies, yielding benefits both at the individual level and for the public health system. This is particularly relevant given the significant challenges faced by the Brazilian healthcare system in managing dementia, including insufficient infrastructure, a limited number of dementia specialists, and disparities in access between the public and private sectors (164).

Research estimates that the waiting list for specialist consultations will increase from approximately 400,000 in 2022 to over 2.2 million by 2040, due to the limited availability of dementia specialists. Brazil currently has only 2.7 specialists per 100,000 inhabitants, compared to an average of 11.27 in G7 countries. This shortage may result in prolonged waiting times for treatment, potentially reaching up to two years on average in the public healthcare sector (164). For this reason, scalable technologies and preventive strategies are essential to help mitigate the challenges faced by the Brazilian healthcare system.

There are several limitations to our study. First, the cross-sectional nature of the dataset prevents the establishment of causal relationships between predictors and outcomes. As such, the results reflect associations only and do not allow for determination of causality. Furthermore, because dementia is a progressive condition, the data used in this study cannot capture temporal dynamics, leading to a loss of information about disease progression or interactions between predictors over time. This limits the generalizability of findings to other time points, particularly in changing contexts. Second, the results were not externally validated with independent cohorts. Third, some variables were self-reported and may be subject to recall bias, affecting the accuracy of certain predictors. Future studies incorporating objective assessments, when feasible, may improve the precision of predictive variables. Finally, although key predictors were included in the model, additional relevant variables might further enhance its predictive performance by providing a more comprehensive understanding.

The strengths of this study include the large sample size, the wide range of collected variables, and the diversity of the population, encompassing individuals of different skin colors, as well as its representativeness of adults aged 50 and older in Brazil. The ELSI-Brazil study incorporated a comprehensive battery of neuropsychological tests designed to provide a broad cognitive assessment. This allowed for the calculation of global cognitive scores using regression-based norms that accounted for the effects of education and age. In contrast, other studies predicting cognitive impairment and dementia have often relied solely on brief screening tools such as the Mini-Mental State Examination (162,165,166). Moreover, ELSI-Brazil included an informant-based cognitive function section, which enabled the assessment of individuals unable to complete the neuropsychological tests, thereby complementing the evaluation and minimizing sample loss.

Future research could benefit from using longitudinal data from ELSI-Brazil, allowing for a more robust analysis of the relationships between predictors and outcomes, as well as the development of predictive models of cognitive status. In addition, it is essential to validate the findings using independent cohorts to assess the generalizability of the results. We also recommend incorporating objective predictors into predictive models to reduce self-report bias. Including other lifestyle-related factors, such as sleep patterns and dietary habits, may further enrich the development of predictive models and provide a more comprehensive understanding of the determinants of dementia.

**4. Conclusion**

According to the results of this study, we observed a prevalence of dementia among Brazilian adults aged 50 years and older that is higher than previously reported in earlier studies. By combining two analytical approaches – Random Forest and multivariable logistic regression – we developed a practical model to estimate the likelihood of dementia, incorporating 10 low-cost and interpretable predictors. The analyses revealed that low educational attainment, older age, low body mass index, reduced handgrip strength, self-reported Black race, physical inactivity, self-reported hearing loss, and the presence of depressive symptoms were consistently associated with higher odds of dementia. Conversely, higher educational attainment, better self-perceived life satisfaction, and currently being employed emerged as protective factors. These predictors encompass sociodemographic, clinical, lifestyle, and psychosocial factors. Their inclusion in the model underscores their relevance in understanding the health determinants related to cognitive status. Moreover, they are accessible within primary care contexts and have the potential to be used in identifying individuals at risk of cognitive and brain vulnerability, especially as such conditions are becoming a leading concern in LMICs, particularly in Brazil, where the health system faces challenges such as inadequate infrastructure, a shortage of dementia specialists, and inequitable access to care. It is therefore essential to focus on integrated approaches that improve early screening in primary care by identifying clinically accessible and relevant factors for preventive strategies. This can enhance resource allocation, both economic and human, while improving the accuracy of timely interventions aimed at preventing or delaying the onset of dementia. Such measures may contribute to addressing the growing burden of dementia in Brazil. Given the cross-sectional nature of this study, we recommend that future research employ longitudinal data and independent cohorts to deepen the temporal understanding of the relationship between predictors and dementia, and to validate the findings observed.

**Conflict of Interest**

The authors declare that the research was conducted in the absence of any commercial or financial relationships that could be construed as a potential conflict of interest.

**Acknowledgments**

I would like to thank my advisors, Dr. João Guilherme and Dr. Tiago Espinola, for their guidance, constant support, and valuable contributions throughout this work. To the Brazilian Ministry of Health for funding the initiative that made the Longitudinal Study of Brazilian Elderly Health (ELSI-Brasil) possible, and to the ELSI-Brasil team for providing access to the dataset used in this study. Special thanks to Dra. Adriana Falcão and Dr. João Garcia for their guidance, advice, and continued support throughout this journey. To Maria Clara Barreto for her operational support.